\documentclass[runningheads]{llncs}
\usepackage{graphicx}
\usepackage{booktabs}
\usepackage{graphicx}
\usepackage{todonotes}
\usepackage{multicol}
\usepackage{algorithmic}
\usepackage{algorithm}
\floatname{algorithm}{Procedure}

\usepackage{adjustbox}
\usepackage{algorithm} 
\usepackage{algorithmic}  
\usepackage[algo2e]{algorithm2e}
\usepackage{csquotes}
\usepackage{adjustbox}
\usepackage{amsfonts}
\usepackage{rotating}

\usepackage{times}
\usepackage{soul}
\usepackage{url}
\usepackage[hidelinks]{hyperref}
\usepackage[utf8]{inputenc}
\usepackage[small]{caption}
\usepackage{graphicx}
\usepackage{amsmath}
\usepackage{booktabs}
\usepackage{algorithm}
\usepackage{algorithmic}
\usepackage{wrapfig}

\usepackage{caption} 
\usepackage{xcolor}
\usepackage{color, colortbl}

\usepackage{etoolbox}

\BeforeBeginEnvironment{figure}{\vskip-4ex}
\AfterEndEnvironment{figure}{\vskip-4ex}

\begin{document}
\title{pH-RL: A reinforcement learning personalization architecture for mobile applications in e-Health}

\title{pH-RL: A personalization architecture to bring reinforcement learning to health practice}

\titlerunning{pH-RL: An RL personalization architecture for mobile applications in e-Health}

\titlerunning{pH-RL: A personalization architecture to bring RL to health practice}

\author{Ali el Hassouni\inst{1,2}\orcidID{000-0003-0919-8861} \and
Mark Hoogendoorn\inst{1}\orcidID{0000-0003-3356-3574} \and 
Marketa Ciharova\inst{3}\orcidID{0000-0002-7131-1549}\and
Annet Kleiboer\inst{3}\orcidID{0000-0001-8040-5697}\and
Khadicha Amarti\inst{3}\orcidID{0000-0002-0586-6407}\and
Vesa Muhonen\inst{2}\orcidID{0000-0002-1812-5288}
Heleen Riper\inst{3}\orcidID{0000-0002-8144-8901}\and
A. E. Eiben\inst{1}\orcidID{0000-0002-3106-4213}}
\authorrunning{el Hassouni et al.}
%
\institute{Vrije Universiteit Amsterdam, Department of CS, Amsterdam, The Netherlands 
\email{\{a.el.hassouni,m.hoogendoorn,g.eiben\}@vu.nl}\and
Mobiquity Inc, Data Science and Analytics, 
Amsterdam, The Netherlands
\email{\{aelhassouni,v.muhonen\}@mobiquityinc.com}\and
Department of Clinical, Neuro- and Developmental Psychology, Amsterdam Public Health Institute, Vrije Universiteit, Amsterdam, Netherlands
\email{\{m.ciharova,a.m.kleiboer,k.amarti,h.riper\}@vu.nl}\\
}
\maketitle              

\begin{abstract}
While reinforcement learning (RL) has proven to be the approach of choice for tackling many complex problems, it remains challenging to develop and deploy RL agents in real-life scenarios successfully. This paper presents pH-RL (\underline{p}ersonalization in e-\underline{H}ealth with \underline{RL}), a general RL architecture for personalization to bring RL to health practice. pH-RL allows for various levels of personalization in health applications and allows for online and batch learning. Furthermore, we provide a general-purpose implementation framework that can be integrated with various healthcare applications. We describe a step-by-step guideline for the successful deployment of RL policies in a mobile application. We implemented our open-source RL architecture and integrated it with the MoodBuster mobile application for mental health to provide messages to increase daily adherence to the online therapeutic modules. We then performed a comprehensive study with human participants over a sustained period. Our experimental results show that the developed policies learn to select appropriate actions consistently using only a few days' worth of data. Furthermore, we empirically demonstrate the stability of the learned policies during the study.  
\end{abstract}

\section{Introduction}

Reinforcement learning (RL) has seen tremendous successes in recent years, principally due to the many breakthroughs made in deep learning (DL) \cite{suttonbarto2nd,mnih2013playing,silver2016mastering,komorowski2018artificial,vinyals2019grandmaster}. The field has witnessed these breakthroughs in high-dimensional control tasks, e.g., complex games Atari and Go and continuous control tasks such as MuJoCo, and openAI gym \cite{DBLP:journals/corr/BrockmanCPSSTZ16}. In many of these tasks e.g., Atari and board games such as Go, Chess, and Shogi, superhuman performance was achieved \cite{silver2016mastering,mnih2013playing,mnih2015humanlevel}. We can attribute these successes to the rise of deep reinforcement learning (DRL) fueled by novel algorithms such as deep Q-network (DQN), the availability of powerful computing hardware, and the nature of the problems at hand that allows one to obtain large samples from the task environment and to perform exploration as one wishes. 

Many practical limitations arise in traditional domains such as healthcare, making these benefits listed above fade away \cite{dulacarnold2019challenges,el2019end,el2018personalization,den2020reinforcement}. Such limitations are the inaccessibility to large samples of data, the unavailability of environments to train and evaluate algorithms in, the limitations on the data caused by privacy laws, and safety concerns (e.g., unsafe actions and exploration), explainability, and legal responsibility \cite{dulacarnold2019challenges,den2020reinforcement}. As a consequence, the applicability of DRL in many practical tasks remains limited. In many practical tasks where RL has been shown to perform well such as advertisement campaign optimisation, there is ample data available, interactions with users are not costly and safety does not play a big role. In healthcare tasks, all these factors play an equally important role. Therefore, there is a need for structural solutions through standardized frameworks and architectures to overcome the abovementioned obstacles and challenges. Consequently, we pose the following research questions and try to answer them with a real-life experiment: \textbf{1) What are the requirements for integrating RL into mobile applications for e-Health? and 2) Can we learn policies quickly that provide personalized interventions?}


This paper presents a general RL architecture (pH-RL) for personalization with the goal of bringing RL to health practice. We propose an RL architecture that allows for adding a personalisation component to applications in healthcare such as mobile applications for mental health. pH-RL allows for different levels of personalisation, namely: pooled (one-fits-all approach), grouped (cluster-level personalisation) and separate (hyper-personalisation on user level). Furthermore, pH-RL allows for online and offline (batch) learning. We describe a step-by-step guideline for the successful deployment of RL policies in a mobile application. 

We implement our open-source RL framework and integrate it with the MoodBuster mobile application for mental health to provide messages to increase daily adherence to the online therapeutic modules. We employ a default control policy approach based on prior knowledge coupled with random exploration in pH-RL. Next, we apply clustering techniques on the traces of states and users' rewards to find an appropriate segmentation of the experiences. Then, we apply online batch RL coupled with exploration driven by the learned policies on each cluster of users. We then perform a comprehensive study with human participants over a sustained period. Our experimental results show that the developed policies learn to select appropriate actions consistently using only a few days' worth of data. Finally, we empirically demonstrate the stability of the learned policies during the study.

\section{Related work}
RL as a solution architecture for real-world problems has seen a significant increase in the last few years ranging from games to advertising and healthcare. This learning paradigm has seen applications in various areas. In this related work section, we discuss RL's real-world applications and architectures in general and Health applications and architectures for personalization in specific.

\newpage
\textbf{Reinforcement Learning Applications} Much effort is put into the use of RL for various applications. A successful application of RL in the last decade was using deep RL to play Atari games \cite{mnih2013playing}. This approach relies on Q-learning with convolutional neural networks to successfully learn control policies for playing Atari games using low-level high-dimensional sensory data as input. These policies surpassed human-level performance in several cases. Also related, \cite{silver2016mastering} combined deep learning with RL to develop policy and value networks that play Go at a superhuman level. These networks learn from extensive amounts of self-play made possible by a well-defined environment with comprehesive rules. In a more recent application, RL has proven to be an effective solution to a real-world autonomous control problem involving the navigation of super-pressure balloon in the stratosphere \cite{bellemare2020autonomous}. This problem is characterized by its complexity, forecast errors, sparse wind measurements, and the need for real-time decision-making. This work uses data augmentation and a self-correcting design to tackle RL issues, usually by imperfect data. As mentioned in Section 1, all these examples do not suffer from the many limitations one encounters with problems in healthcare.

\textbf{Reinforcement Learning for Clinical Applications} A literature review has shown that the number of applications of RL has been increasing \cite{den2020reinforcement}. Applications in healthcare range from treating patients with Sepsis at the Intensive Care Unit to sending personalized messages in e-Health mobile applications. 
There is strong evidence that suggests that current practices at the ICU are not optimal while the best treatment strategy remains unknown. (Komorowski et al.) used RL to develop policies that are on average more reliable than human clinicians \cite{komorowski2018artificial}. This approach provides individualized treatment decisions that are interpretable by clinicians. Similarly, \cite{saria2018individualized} used RL to develop policies for individualized treatment strategies to correct hypotension in Sepsis. More recent work showed that these developed policies for optimizing hemodynamic treatment for critically ill patients with Sepsis are transferable across different patient populations \cite{roggeveen2020transatlantic}. Furthermore, this work proposes an in-depth inspection approach for clinical interpretability. In a strictly regulated area such as healthcare, these example are considered very innovative. However, structural solutions are needed. Our pH-RL standardized architecture and the corresponding generic framework allow us to bring online and offline RL for personalization to health practice.

\textbf{Reinforcement Learning for Personalization in e-Health Applications} More and more real-world applications using RL for personalization in e-Health are found in literature \cite{den2020reinforcement}. Work by \cite{el2018personalization} focuses on developing RL policies coupled with clustering techniques for personalizing health interventions. They show that clustering using traces of states and reward and developing policies based on these clusters leads to improved personalization levels while speeding-up the learning time of the approach. In a later work, \cite{el2019end} demonstrates that this approach leads to improved personalization levels when applied on state representations consisting of raw sensor data obtained from mobile apps. Similarly, k-means clustering and RL were combined to develop policies across similar users for the purpose of learning better policies \cite{zhu2018group}. Clustering methods were also used to effectively learn personalized RL policies in health and wellbeing \cite{grua2018exploring}.

\newpage

\textbf{Personalization Architectures for e-Health Mobile Applications} A wealth of mobile apps exist that support people in their daily lives. We can use these apps for mental coaching, health interventions, fitness apps, and various other purposes. Although these applications can take various types of information into account, such as location, and historical behavior, they still rely on rule-based approaches and do not achieve high effectiveness and efficaciousness of treatment. Reinforcement learning-driven personalization has proven to be a practical approach for many health settings, including e-health and m-health \cite{den2020reinforcement}. 
Furthermore, clinical support systems could rely on the same techniques to achieve high effectiveness and officiousness of treatments. Research shows a lack of publications that propose RL architectures for personalization. \cite{grua2020reference} presents a reference architecture that enables self-adaptation of mobile apps for e-Health. Although this architecture proposed a non-rule-based approach for self-adaptation, it does not specify machine learning techniques to achieve personalization. This architecture relies on MAPE loops that operate at different levels of granularity meant for different purposes. (Hoffman et al.) proposed a research framework for distributed RL (Acme) \cite{hoffman2020acme}. This framework aims at simplifying the process of developing RL algorithms in academia and industry.

\section{pH-RL - An RL personalization architecture for health practice}

This section introduces our reinforcement learning personalization architecture for mobile applications in e-Health. We start by framing the problem definition. Then we introduce our framework for personalization with RL. Finally, we present our pH-RL framework for personalization in e-Health.

\subsection{The RL architecture for personalisation}

\begin{figure*}[]
\includegraphics[angle=0,origin=c,width=12cm, height=5.5cm, trim=0cm 0cm 0.5cm 3.0cm]{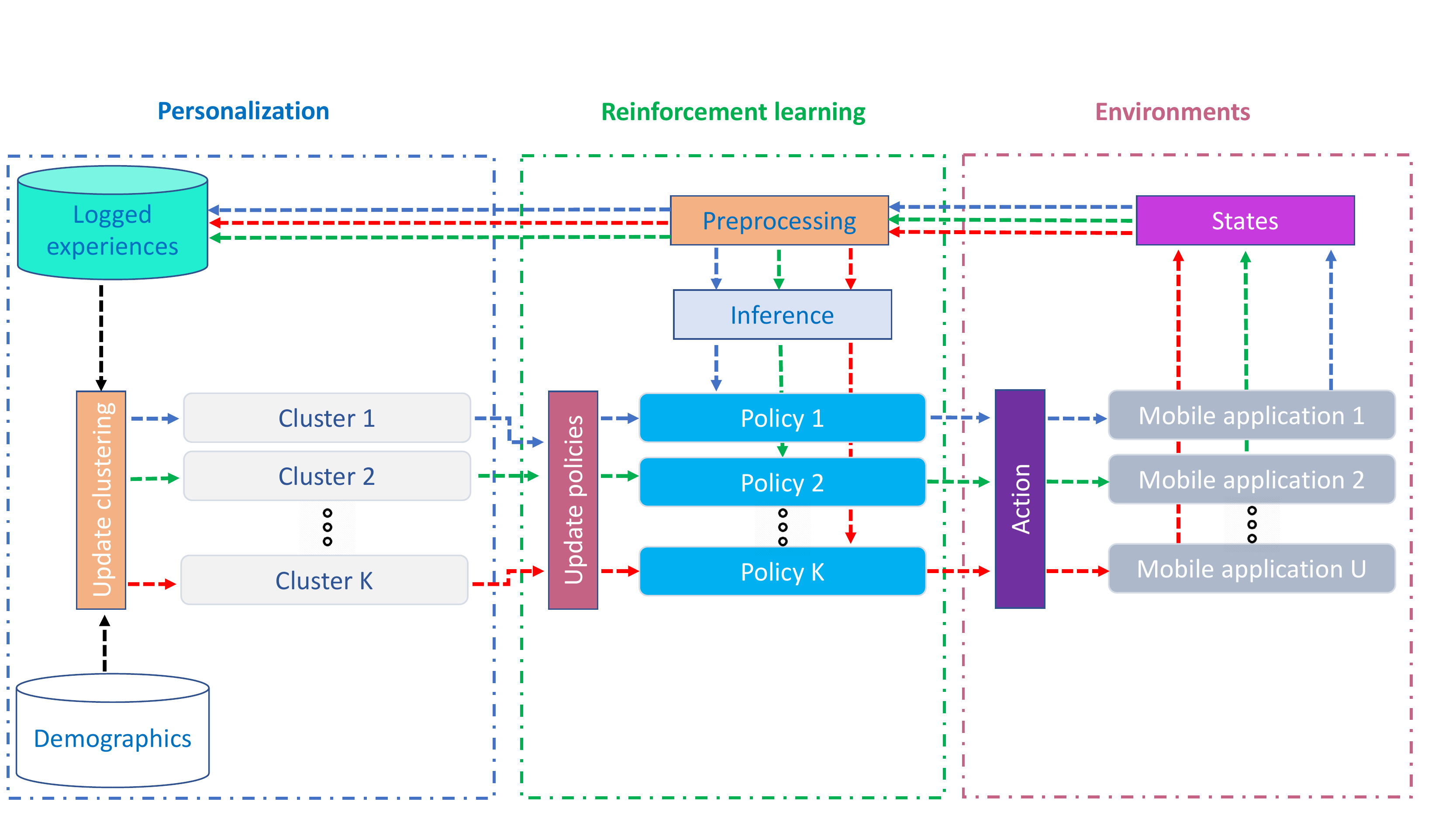}
\centering
\caption{pH-RL: A reinforcement learning personalization architecture for mobile applications in e-Health} 
\label{overview_system}
\end{figure*}

Figure ~\ref{overview_system} shows the pH-RL framework for personalization of interventions in e-Health applications using RL. This section starts with a definition of the system's different components and the environment it is interacting with to learn personalized policies. 

\subsubsection{Users of the mobile applications} Users are the people that utilize the e-Health application to get help achieving a specific health goal. These users own a smart device that can run the e-Health application. They install the e-Health applications and continue to sign-up to be able to use the application. We ask these users to provide information about their demographics, preferences, and goals to form the initial clusters. After the sign-up phase, the users start generating data used by the pH-RL framework to personalize the interventions further. The e-Health applications send these interventions to help the users achieve their goals. 

\subsubsection{Smart device} This device can be any computer that can host applications and have an interface to interact with the users. Mobile phones and smartwatches are the most natural types of smart devices used to host e-Health applications. These devices contain sensors that allow the pH-RL framework to obtain potentially very granular data about the users' behavior continuously. We can use this information to infer contextual information about the users' state without having to ask them to provide this information at all times explicitly. Additionally, smart devices contain interfaces in the form of a screen or voice that allows the users to receive interventions from the e-Health applications and provide feedback to the application. 

\subsubsection{State} A state contains information about the user at a certain point in time $t$. We capture this information through the e-Health application. To model a real-world problem as an RL problem, one has to design a Markovian state representation. In pH-RL, a state $s \in S$ is made up of features represented by the feature vector representation $\vec{\psi}(s) = \langle\psi_1(s),\psi_2(s),\ldots,\psi_n(s)\rangle^\top$. A feature can indicate the occurrence of a certain activity, the number of times a user performed a certain activity during a specified period, or any other feature that provides useful information about the user's behavior. In pH-RL, we perform binning to transform any continuous feature into a discrete representation.

\subsubsection{Interventions} E-Health applications running pH-RL are used by people to help them with a health-related goal. The application gathers data from the user's smart device and sends it to our pH-RL framework, which returns an action $a \in A$ presented to the user in the form of interventions. The most natural way of providing a user with an intervention is via notifications of mobile applications or smartwatches. The main goal in pH-RL is to provide users with interventions that are relevant at the right moment. This hypothesis is that personalized interventions will lead to a higher long-term adherence level to the user's goal. 

\newpage
\subsubsection{Rewards} The pH-RL framework gathers data from the users through the e-Health application. Behavioral change data from the users about the intervention (either direct or indirect) is used as a feedback mechanism used by the RL algorithm in pH-RL as the reward signal to learn personalized policies. For instance, the direct reward can be users providing ratings about their moods a few times per day in a mental health application. Indirect rewards can be the amount of activity measured after receiving the intervention. Rewards play an essential role in pH-RL and should be defined carefully to represent the problem at hand well.


\subsection{Preliminaries and Problem Statement}

\subsubsection{Reinforcement Learning}
We model a real-world problem $M$ as a Markov Decision Process ($MDP$). Therefore, we consider the task $M$ to be Markovian and therefore it can be modelled using RL algorithms. In this section we borrow the RL terminology from \cite{el2018personalization}. We define $M$ to be $\langle S$, $A$, $T$, $R \rangle$ where $S$ is a finite state space, and $A$ the set of actions that can be selected at each time step $t$. There exists a probabilistic transition function $T$ :: $S \times A \times S \rightarrow$ [0,1] over the states in $S$. When at time $t$ in current state $s$ an action $a$ is selected, a transition is made to a next state $s' \in S$ at $t+1$. $R :: S \times A \rightarrow \mathbb{R}$ is the reward function and outputs a scalar $r = R(s, a)$ to each combination of state  $s \in S$ and action $a \in A$. The feature vector representation $\vec{\psi}(s) = \langle\psi_1(s),\psi_2(s),\ldots,\psi_n(s)\rangle^\top$ defines the features that form the states $s \in S$. Our aim when modelling task $M$ as an RL problem is to learn a policy $\pi$. With the policy $\pi :: S \rightarrow A$ we can determine which action $a \in A$ to take in a state $s \in S$ such that the long term expected cumulative reward is maximized. Selecting action $a = \pi(s)$ will result in a transition to state $s^\prime$. Here, a reward $r = R(s,a)$ will be obtained. Taking an action while in a state and transitioning to a new state forms an an experience $\langle s, a, r, s^\prime \rangle$. A trace is a sequence of experiences in a particular order. Denote a trace by $\zeta$ : $\langle s, a, r, s^\prime, a^\prime, r^\prime, s^{\prime\prime}, a^{\prime\prime}, r^{\prime\prime}, \ldots \rangle$. Multiple transitions over time result in multiple experiences. The combination of experiences over time form a data set $Z \in \langle \zeta_1,\ldots \zeta_k\rangle$.

Reinforcement learning aims at learning the best policy $\pi^*$ out of all possible policies $\Pi$ :: $S \times A \rightarrow$ [0,1]. $\pi^*$ selects actions with the aim of maximizing the the sum of future rewards (at any time $t$). We assign a value for taking action $a \in A$ in state $s$ of policy $\pi$ $\pi(s)=a$ as follows:

\vspace{-15pt}
\begin{equation}
	\label{eq:policyeval}
	Q^\pi(s,a) = E_\pi \large\{ \sum_{k=0}^{K} \gamma^k r^{t+k+1} | s^t=s, a^t=a \large\}
\end{equation}
\vspace{-5pt}

Here $\gamma$ is the discount factor that gives weights to future rewards. $s^t$ and $a^t$ define the states and actions at time $t$. Denote $Q(s,a)$ as the expected long-term value of state $s$ after taking action $a$. If we select the best action $a$ in each possible state $s \in S$, a policy can be derived from the $Q$-function, i.e. 

\vspace{-7pt}

\begin{equation}
	\label{eq:argmaxQ}
	\pi^\prime(s) = \arg\max_{a \in A} Q^{\pi}(s,a), \ \forall s \in S
\end{equation}

\newpage
\subsubsection{Cluster-Based Policy Improvement}

One-fits-all approaches are based on the assumption that users belong to one group, and therefore, one policy is learned across all these users. This approach has been shown to perform sub-optimally in e-Health applications because people, in general, have different preferences and are characterized by non-identical transition and reward functions \cite{el2018personalization}. 

In the pH-RL framework, we mitigate this issue and propose to group users with similar behavior using clustering techniques \cite{el2018personalization,el2019end}. Clustering algorithms such as K-medoids and K-means have been shown to perform well on similar problems in e-Health \cite{el2018personalization,el2019end,grua2018exploring}. We compare the behavioral traces of users consisting of states and rewards, and we use the Dynamic Time Warping (DTW) algorithm to calculate the distance between two users. Several other distance metrics, such as the Euclidean distance, have been explored in e-Health literature. Although metrics are considered valid approaches for calculating distances between two traces, DTW is more accurate because it measures the similarity between two different users' traces. It does so by finding the optimal match between two potentially similar traces that are out of phase where the Euclidean distance would have found these two traces to be very different. We define the traces of $u$ consisting of states and rewards as:
$\langle s_u, r, s_u^\prime, r^\prime, s_u^{\prime\prime}, r^{\prime\prime}, \ldots \rangle$.

Define the group of users to be targeted by our framework as $U$ and $\Sigma^U$ as all the traces generated by these users. The experiences (excluding the actions) generated by user $i$ during day $d$ are defined as $\Sigma^{\prime u{_{i,d}}}$. We calculated the similarity between two users $u_1$ and $u_2$ as: 

\vspace{-15pt}

\begin{equation}
	\label{eq:dtw}
	S_{DTW}(u_1,u_2) = \sum_{d=0}^{D} dtw(\Sigma^{u{_{1,d}}},\Sigma^{u{_{2,d}}}).
\end{equation}

We apply a clustering method to obtain clusters $k$ of users based on their traces where $k$ and $\Sigma^U_1,\ldots,\Sigma^U_k$ is a partitioning of $\Sigma^U$, and let $U_1, \ldots U_k$ be the partitioning of $U$ \cite{el2018personalization}. In a one-fits-all approach, we would utilize all experiences of $U$ to learn one $Q$-function. In pH-RL, we learn a distinct $Q$-function $Q_{\Sigma^U_i}$ and policy $\pi_{\Sigma^U_i}$ for each user set $U_i$ based on all the traces in $\Sigma^U_i$. Note that these steps are done in addition to our previous setup, which allows for a comparison between a policy for $U$ and subgroup policies. 

\begin{table}[ht]
\centering 
\begin{tabular}{l l } 
\hline\hline 
Feature & Definition \\ [0.5ex] 
\hline 
Day Part & A numerical encoding for part of day (0: morning, 1: afternoon, and 2: evening).   \\ 
Number Rating & The cumulative number of ratings inputted by the user.  \\
Highest Rating & The highest rating inputted by a user during the current day.  \\
Lowest Rating & The lowest rating inputted by a user during the current day.  \\
Median Rating & The median rating inputted by a user during the current day.  \\
SD Rating &  The standard deviation of the ratings inputted by a user during the current day.\\
Number Low Rating &  The number of low (1 and 2) ratings inputted by a user during the current day.\\
Number Medium Rating &  The number of medium (3, 4, and 5) ratings inputted by a user during the current day.\\
Number High Rating &  The number of high (6 and 7) ratings inputted by a user during the current day.\\
Number message Received &  The number of messages received by a user during the current day.\\
Number Message Read &  The number of messages read by a user during the current day.\\
Read All Message &  Indicator if a user reads all messages during the current day.\\
[1ex] 
\hline 
\end{tabular}
\caption{State features, actions and reward definitions.} 
\label{table:features} 
\end{table}
\vspace{-4mm}
\subsection{Framework implementation and algorithm setup}
In this section, we discuss the framework implementation and our algorithmic setup. Furthermore, we discuss in detail our design choices for the state, actions, and rewards. Our proposed pH-RL architecture can be applied across many personalization tasks in mobile applications. In this section, we demonstrate an instance implementation of pH-RL for mental health using the MoodBuster platform.

\subsubsection{State}
We designed features (i.e. $\vec{\psi}(s_u)$) to represent the state of user. Table \ref{table:features} shows an overview of the features. These feature were designed to capture the behaviour of the user on the mobile application and their mood ratings.

\newpage
\subsubsection{Action}
We use four actions that the policy can select. Action $0$ represents: "send no message",  action $1$ represents sending an action of type: "encouraging", action $2$ represents sending an action of type: "informing", and action $3$ represents sending an action of type: "affirming". Once the action is selected, we further decide from which sub-group of actions to select based on the user's mood. All messages and categories, including the splits based on mood, can be found in the appendix. To make sure we do not send the same message multiple times during a day, we randomly select a message from the set of possible messages that were not previously selected during the same day.

\subsubsection{Reward} 
Adherence can be measured by how often users are using the MoodBuster application. Therefore, the reward function is a combination of two components weighted equally. The first component measures the fraction of messages received during a day up until the current daypart. The second component measures the number of ratings inputted by a user during a day until the current day. 

\subsubsection{Least Squares Policy Iteration (LSPI)}
We perform training using (batch) online learning with the LSPI algorithm because of its ability to generalize well on relatively small datasets. Every time the policies for the different clusters of users are updated with a new batch of data, we export a policy for inference of actions. We use the exact basis function transformation of our features by first binning each of the features into four bins, increasing our features by a factor of 4. We use a policy with the LSTDQ Solver, a discount factor of $0.95$, an exploration rate of $0.1$, the tie-breaking strategy first wins, max iterations of 25, and stopping criterion $\epsilon$ of $0.00001$. These hyper-parameters are based on various experiments from earlier work in this area \cite{el2018personalization,el2019end}.
                  
\newpage
\subsubsection{Technical implementation of the pH-RL system}
We utilize Amazon Web Services to run our pH-RL system. Our setup consists of an S3 bucket to store RL experiences safely and securely. pH-RL is implemented as an open-source Python package \cite{ali_el_hassouni_2021_4628543}. The code is deployed on an AWS EC2 instance. The scripts for performing batch training (once a day), clustering (once during the entire experiment), inference, or sending a message (three times every day) are run with a time-based job scheduler (cronjob). We make secure connections with the restful API of MoodBuster to make read calls to retrieve data and post calls to send messages. 

\section{Real-world performance evaluation}

To evaluate the proposed pH-RL framework, we conducted a real-world experiment with a mobile application for mental health. This experiment's main aim is to demonstrate the feasibility of applying the pH-RL framework for personalization in e-Health applications and provide easy-to-follow guidelines for successful integration and deployment of RL models in real-world applications. We integrated our pH-RL framework into the MoodBuster platform to provide personalized messages and answer our first research question.

\subsection{Personalized motivational feedback messages}
We designed a real-world experiment to improve adherence to an online course for low mood with personalized motivational feedback messages. The pH-RL framework is used to select the most appropriate messages to send to a user to maximize adherence to the course. We consider three categories of messages based on existing research \cite{schwebel2018using}. Informing messages aim at providing informative messages to help the user understand the MoodBuster platform and the online course. Encouraging messages try to encourage users to perform a specific action. Affirming messages offer emotional support or encouragement. We further split the encouraging and affirming categories based on the user's mood into the following three groups: positive, neutral mood, negative, neutral mood, and mood unavailable. 

\subsection{MoodBuster}
MoodBuster is a research platform that has been developed to treat psychological complaints online. Treatments on MoodBuster take place in connection with research projects. The platform gives access to two types of applications: a web application for patients, a web portal for practitioners, and a mobile application that can measure information such as the Ecological Momentary Assessment (mood and user state measurements) \cite{mikus2018predicting,provoost2020improving}. The platform can be used to treat patients as part of a guided online treatment or as a prevention and self-help tool. Furthermore, MoodBuster can be used as part of guided online treatments or blended treatment (face-to-face therapy combined with online treatment. In this work, we use the cognitive behavioral therapy treatment for depression on the MoodBuster platform. This treatment consists of 6 online modules containing readable and watchable material to guide the user through the module to perform exercises and assignments. The cognitive-behavioral therapy treatment for depression helps the user better understand depression, stimulate positive thinking, stimulate behavioral change through enjoyable activities and physical exercise.

\subsection{Participants}
We conducted this experiment with 30 participants from our research departments at the Vrije Universiteit Amsterdam, students, and friends. All these participants were not selected based on having depression. Furthermore, we informed these users that we performed this study to test our pH-RL framework for personalization. Henceforth, they should use this application with the intent that it is part of a test case.

\subsection{Setup of runs}
We train our policies using batch learning. We set the intervention moments when the application sends messages at 10:00, 14:00, and 21:00. We perform random exploration during the first week of the experiment by sending three random interventions a day. After week one, we perform clustering to find our clusters and train our policies using all available data from the start of the experiment. Phase 2 of the experiment runs for an additional two weeks using actions from the learned policies. We update our policies in batch at the end of every day at 23:59. 

\section{Results}

In this section, we present the results related to the experiment laid out in section 4. We experimented with 27 participants to test our integration of the pH-RL framework for personalization in e-Health with the MoodBuster mobile application for depression. We instructed the participants to use and interact with the application to generate data 
We do not assume any of the participants to have any symptoms of depression, and therefore we do not expect the app to lead to any significant changes in the participants' mood. Our main aim is to present, implement and test the pH-RL architecture in an e-Health setting and demonstrate that we can learn policies quickly that provide personalized interventions.

\subsubsection{Inactive participants}
During the experiment, a fraction of all participants were entirely dormant and showed no activity. Inactive users reflect actual real-life usage stats of mobile applications. The experiences of these users are still included in the data set provided used to develop the models. During the whole experiment, 6 participants showed barely any activity and were excluded (entered a rating or read a message at least once for maximum of two days). Furthermore, an additional eight users became inactive during weeks 2 and 3 of the experiment. During the data analysis, we reported different results that include and exclude these users. In this experiment users becoming inactive is not the results of the interaction with the application. 

\subsection{Exploration phase}
Developing RL policies requires a data set consisting of many experiences. At the start of our experiment, we lack such data. As a solution, we implemented a default policy to be used during the experiment's initial phase for one week. During this phase, random exploration was applied, resulting in the data presented in figure 2. From the figure on the right, we can see that around 70\% of the messages were read throughout the first week. Furthermore, we observed that around 75\% of the users rated their mood at least once per day during the first week with a drop on Friday and Saturday.

\begin{figure}[H]
    \centering
    \begin{adjustbox}{max width=\linewidth}
    \begin{tabular}{c c c c}
    \includegraphics[width=\textwidth]{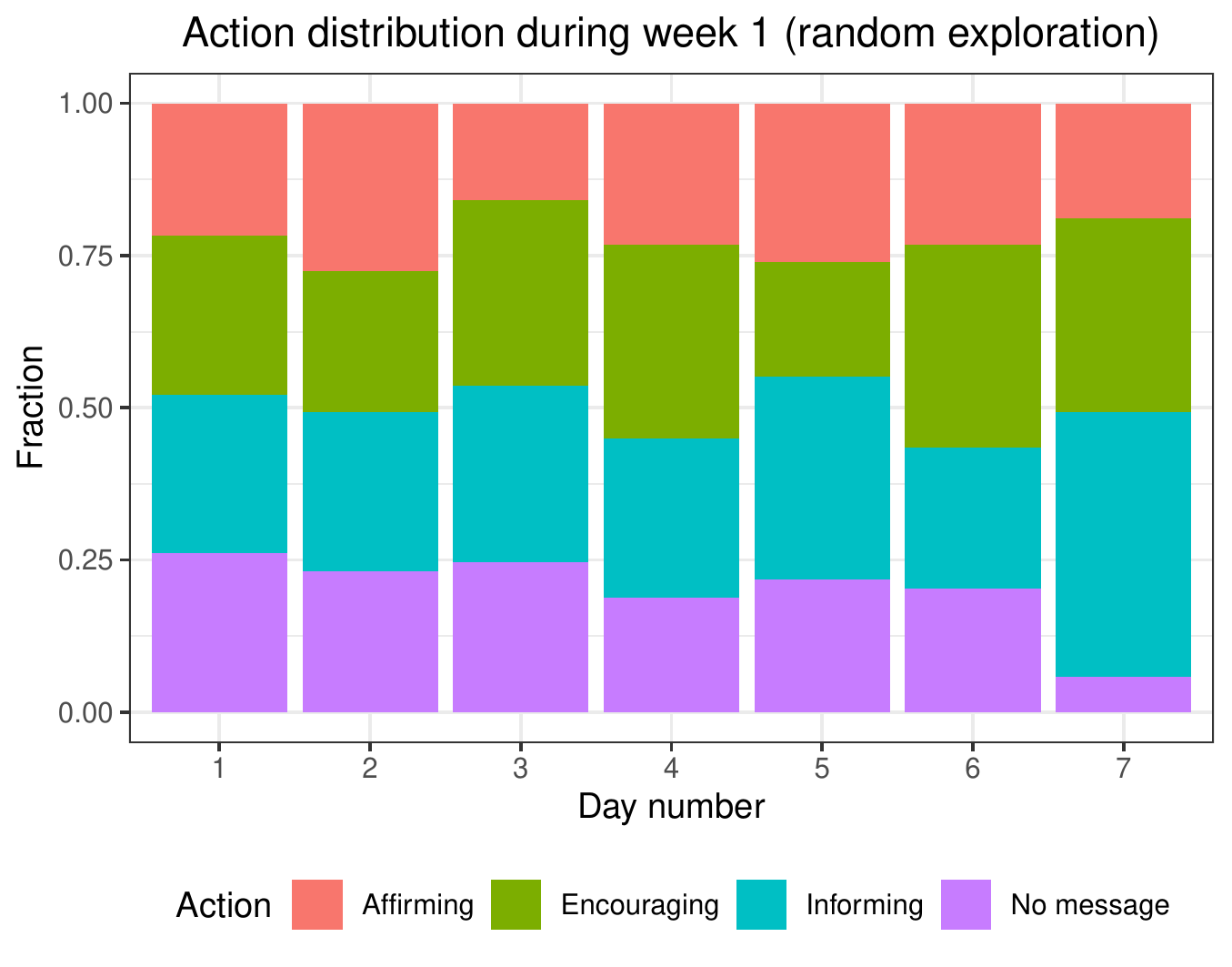}&
    \includegraphics[width=\textwidth]{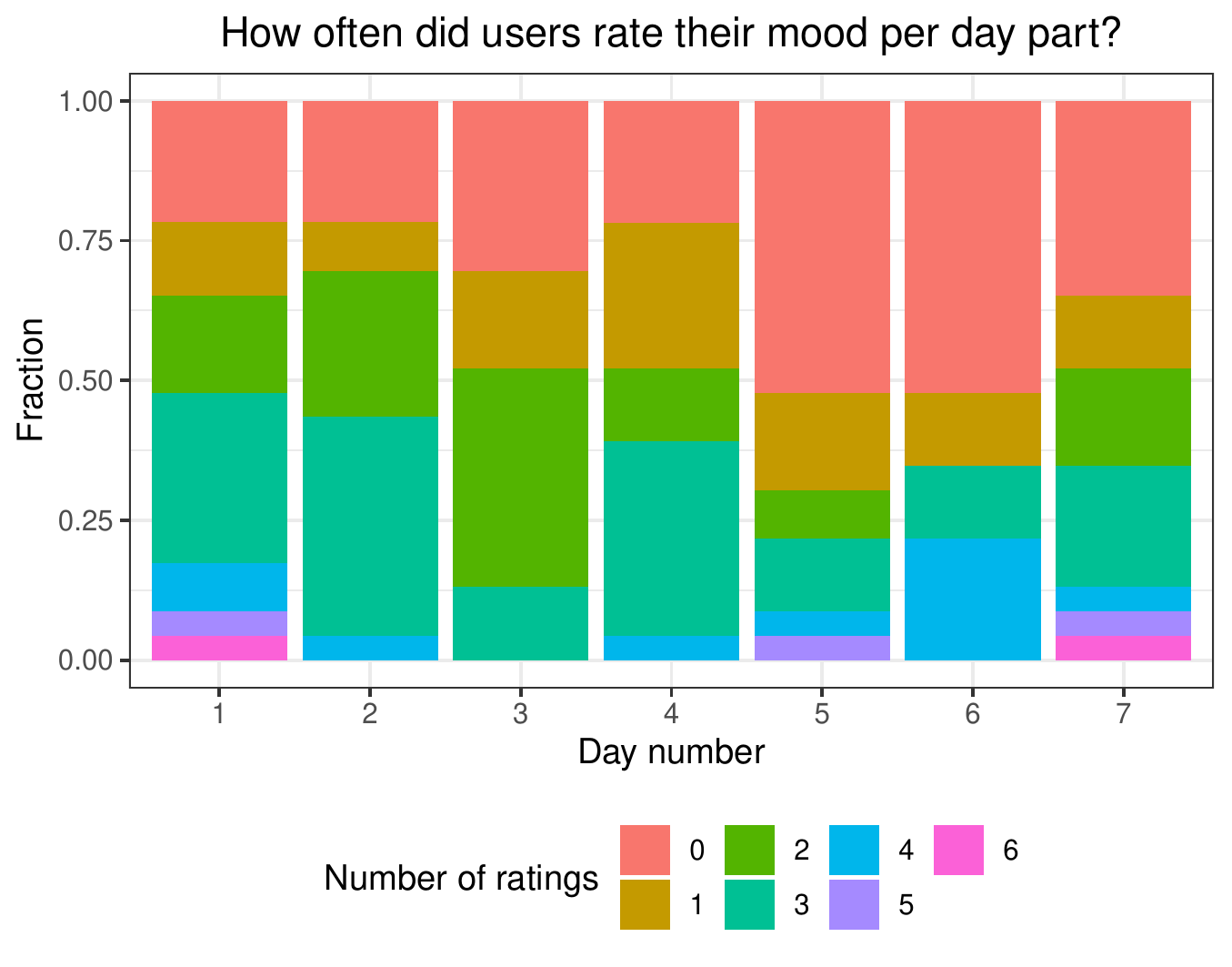}&
    \includegraphics[width=\textwidth]{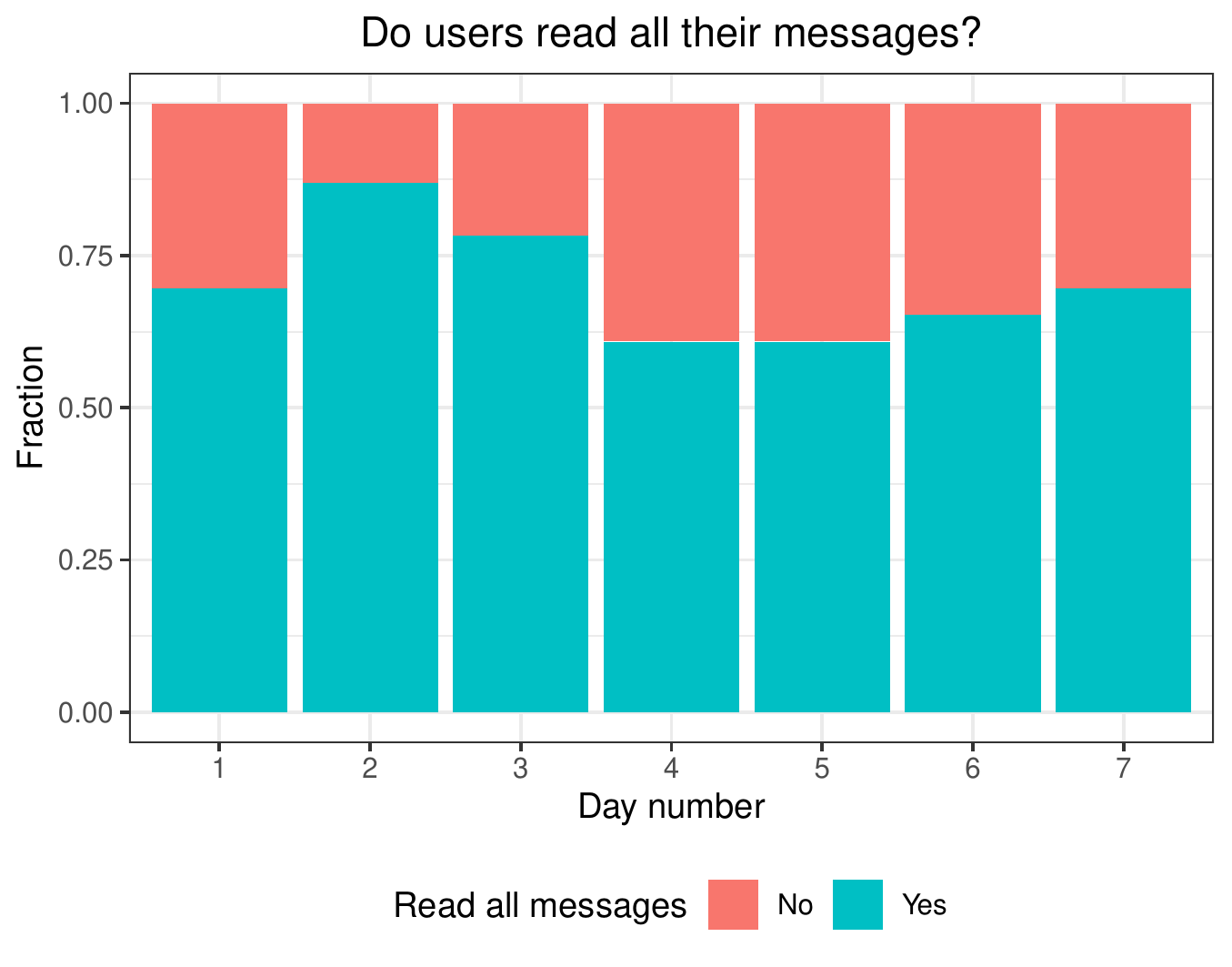}&
    \end{tabular}
    \end{adjustbox}
    \caption{Week 1 action distributions (left), number of ratings per day (middle) and fraction of messages read (right).}
\label{figure3}
\end{figure}

\subsection{Learning phase}
After phase 1, we trained the first policy and started using it. We update the policy with new data at the end of every day. Here we do include data from all 8 participants that became inactive after week 1. Using the obtained results we answer the second research question in this section. Figure 3 (middle) shows how often users rated their mood per day. On average, at least 45\% of the participants rated their mood at least once per day.  Finally, we see from figure 3 (right) that around 50\% of the users read all their messages per day. Figure 3 (left) shows the distribution of actions for all dayparts. We can see that the policy started favoring "Informing" messages around 50\% of the time. As we move forward in time, the policy starts favoring encouraging messages more often. This is in line with what we hypothesized based on existing research \cite{bailoni2016healthy,schwebel2018using}. On day 7 of the second week, the policy decides not to send a message in around 40\% of the cases.  Overall the policy favors the actions in the following order: "encouraging", "informing", "affirming", and "sending no message".

\begin{figure}[H]
    \centering
    \begin{adjustbox}{max width=\linewidth}
    \begin{tabular}{c c c c}
    \includegraphics[width=\textwidth]{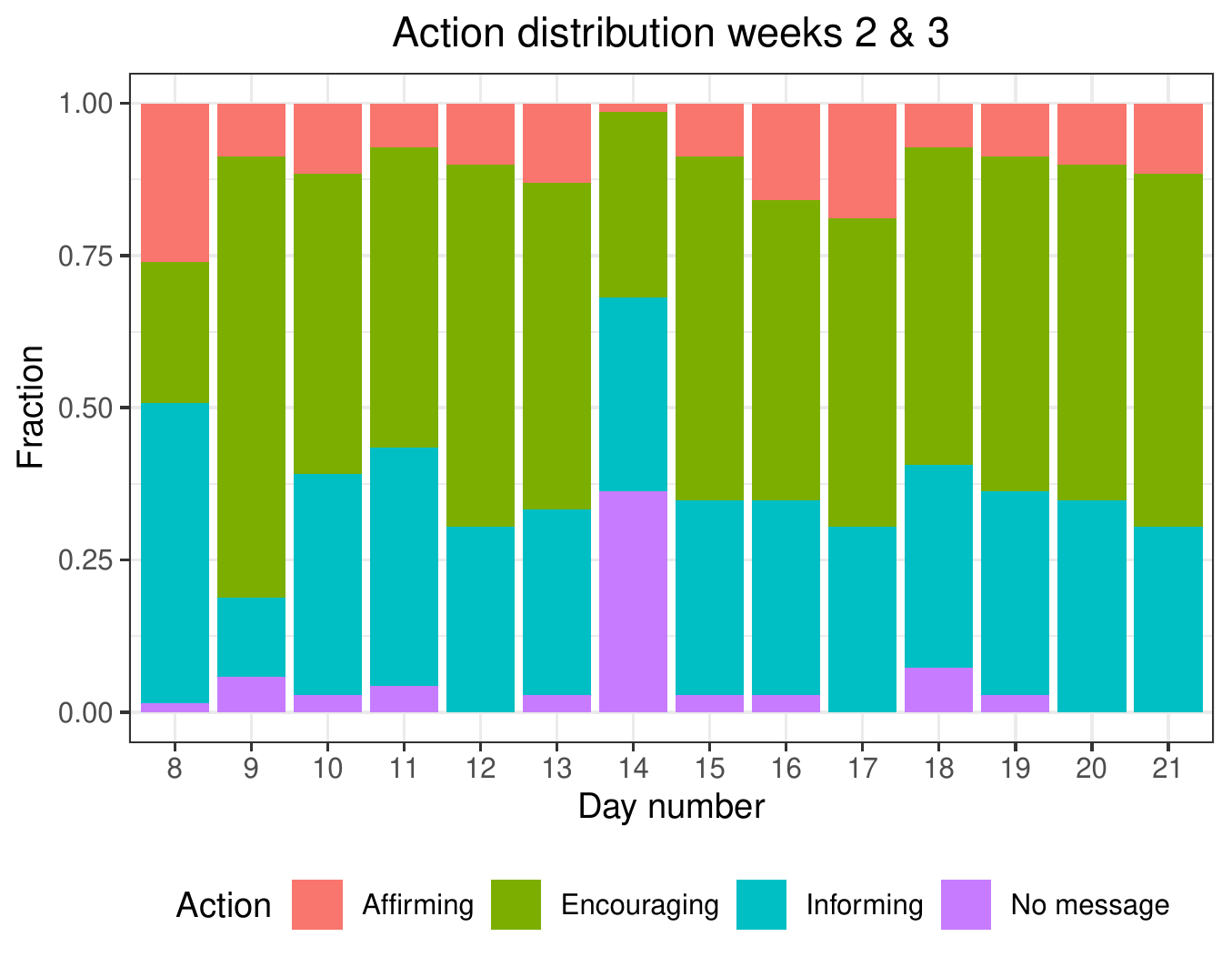}&
    \includegraphics[width=\textwidth]{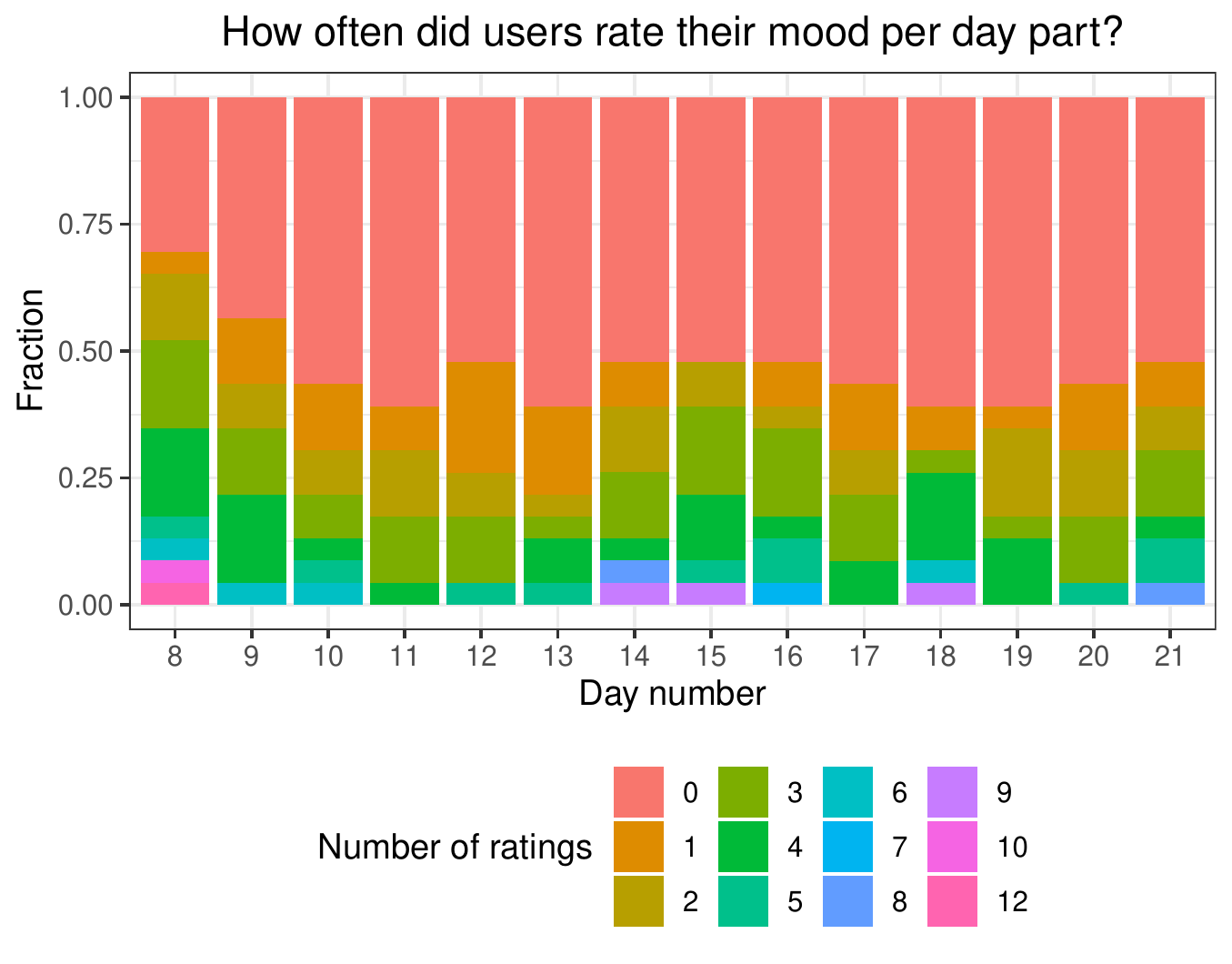}&
    \includegraphics[width=\textwidth]{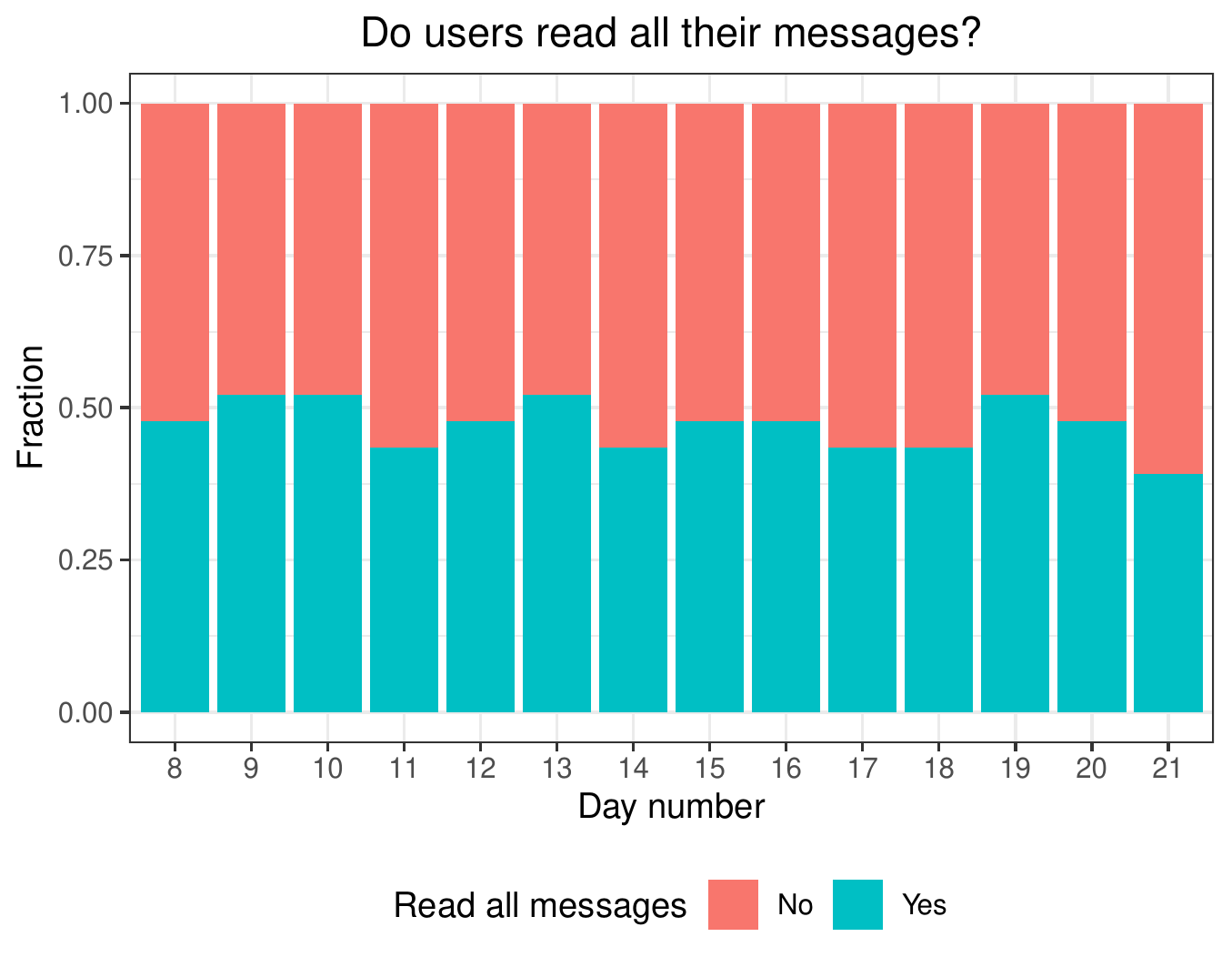}&
    \end{tabular}
    \end{adjustbox}
    \caption{Week 2 and 3 action distributions (left), number of ratings per day (middle) and fraction of messages read (right).}
\label{figure3}
\end{figure}

To further understand how the policy is personalizing towards specific attributes in the state space, we created the visualization that shows the action distributions of the policies per daypart in figure 4. In figure 4 (left), we can see that the policy started favoring "informing" messages in the morning but quickly changed to a strategy with "encouraging" messages. In the afternoon, we see similar behavior with more "affirming" messages. On day 14 of the experiment, the policy does not send any messages in the afternoon. Finally, the evening strategy is to mostly send "encouraging" messages during the first two days and then switch to a strategy dominated by "informing" messages in the evening. These findings are in line with existing research \cite{schwebel2018using}.

\begin{table}
\centering
\small
\begin{tabular}{c|c|c|c|c|c} 
\hline
\multicolumn{1}{c}{} & \multicolumn{1}{c}{\textbf{Week 1} } & \multicolumn{1}{c}{\textbf{Week 2} } & \multicolumn{1}{c}{\textbf{Week 2*} } & \multicolumn{1}{c}{\textbf{Week 3} } & \textbf{Week 3*}       \\ 
\hline
\textbf{All day parts}     & 0.98 ± 0.79                          & 0.83 ± 1.02                          & \textbf{1.20 ± 1.06}                  & 0.77 ± 0.98                          & 1.16 ± 1.01            \\ 
\hline
\textbf{Morning}     & 0.66 ± 0.57                          & 0.62 ± 0.78                          & \textbf{0.87 ± 0.81}                  & 0.54 ± 0.64                          & 0.81 ± 0.65            \\ 
\hline
\textbf{Afternoon}   & 1.1 ± 0.80                           & 0.87 ± 1.0                           & 0.87 ± 0.81                           & 0.83 ± 1.01                          & \textbf{1.25 ± 1.03}   \\ 
\hline
\textbf{Evening}     & 1.2 ± 0.87                           & 1.00 ± 1.19                          & \textbf{1.46 ± 1.22}                  & 0.93 ± 1.16                          & 1.40 ± 1.19            \\
\hline
\end{tabular}

\caption{The average reward per user per daypart for weeks 1, 2, and 3. Week 2* and Week 3* are with inactive users excluded. }
\label{table:results}
\end{table}

\begin{figure}[H]
    \centering
    \begin{adjustbox}{max width=\linewidth}
    \begin{tabular}{c c c c}    
    \includegraphics[width=\textwidth]{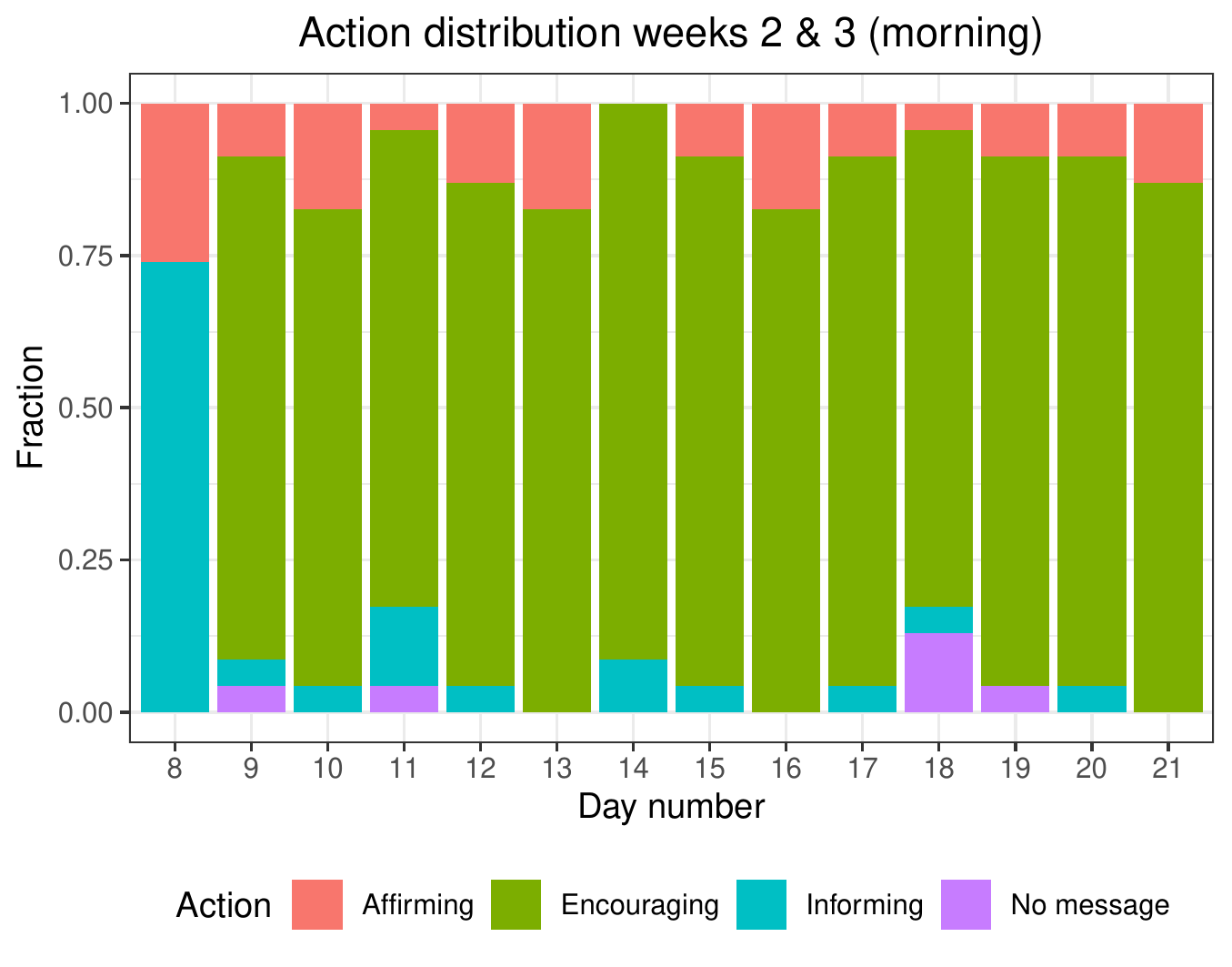}&
    \includegraphics[width=\textwidth]{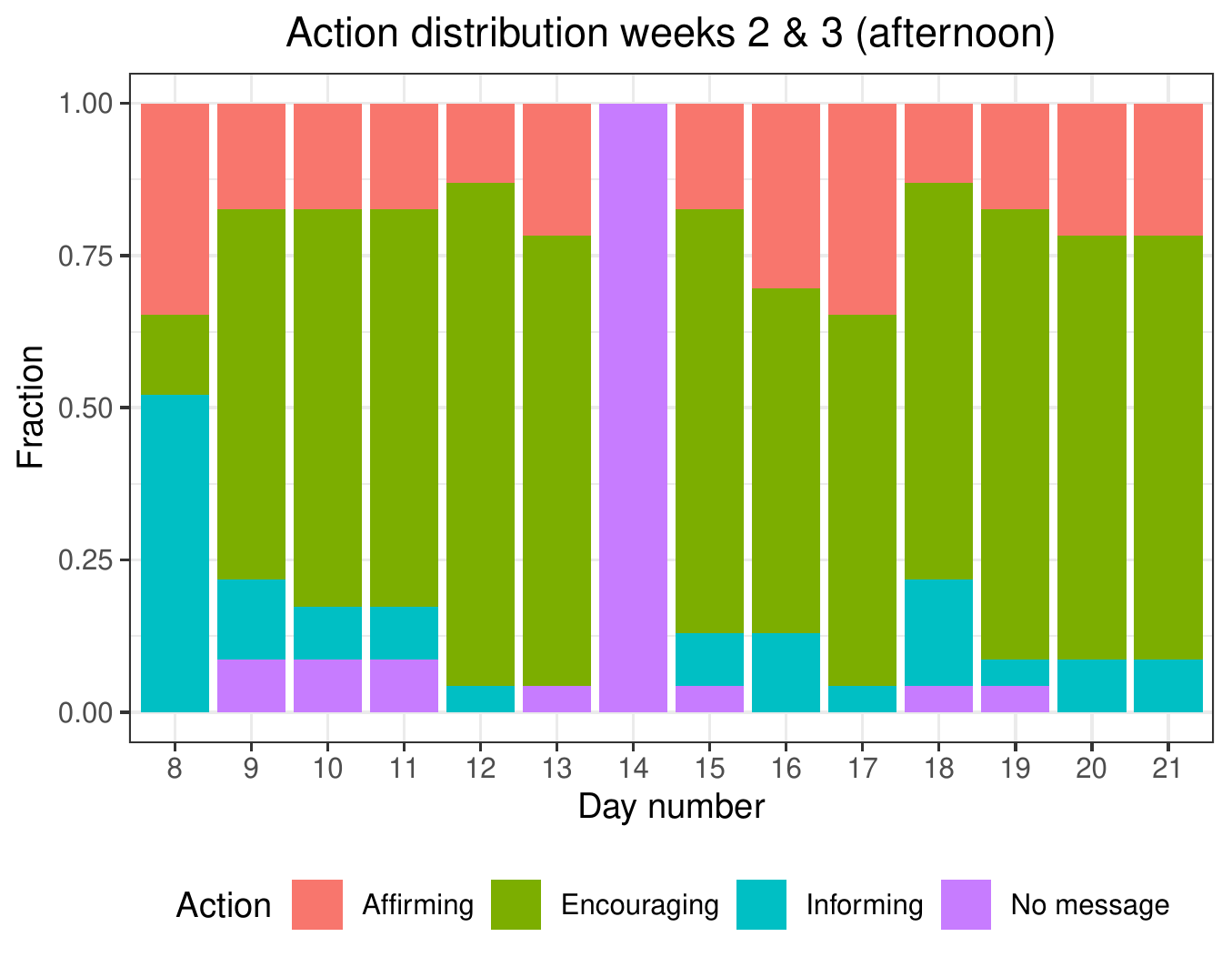}&
    \includegraphics[width=\textwidth]{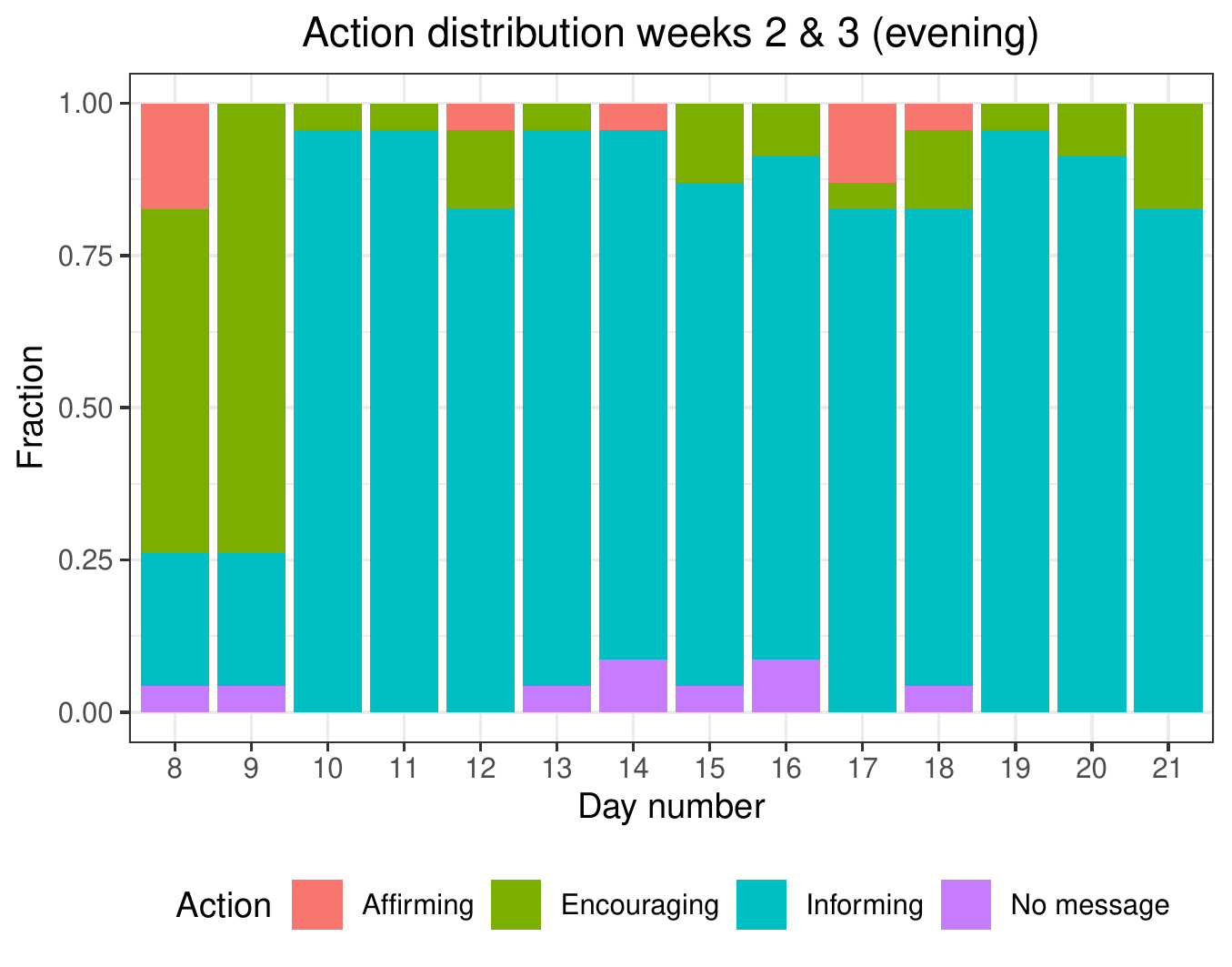}&
    \end{tabular}
    \end{adjustbox}
    \caption{Week 2 and 3 action distributions morning (left), afternoon (middle) and evening (right).}
\label{figure3}
\end{figure}

\newpage

\subsubsection{Rewards}
Figure 5 and Table \ref{table:results} show the observed rewards during weeks 1, 2, and 3. Furthermore, we also consider statistics and make comparisons after excluding all inactive users after week 1. We can see that the average reward per day part per user drops as time proceeds. One of the main reasons that a certain number of users stopped using the app after week 1 was because some were instructed to do so. Furthermore, other users stopped using the app once they felt they have performed enough testing. When we exclude users that dropped out, the average reward increases again. Furthermore, we observe that users obtain higher rewards consistently in the evening. Our hypothesis has to do with the fact that people have more time to check their phones during these moments of the day. From Figure 5, we also see that the maximum reward per user per day part increased after week 1. 

\begin{figure}[H]
    \centering
    \begin{adjustbox}{max width=\linewidth}
    \begin{tabular}{c c c c}    
    \includegraphics[width=\textwidth]{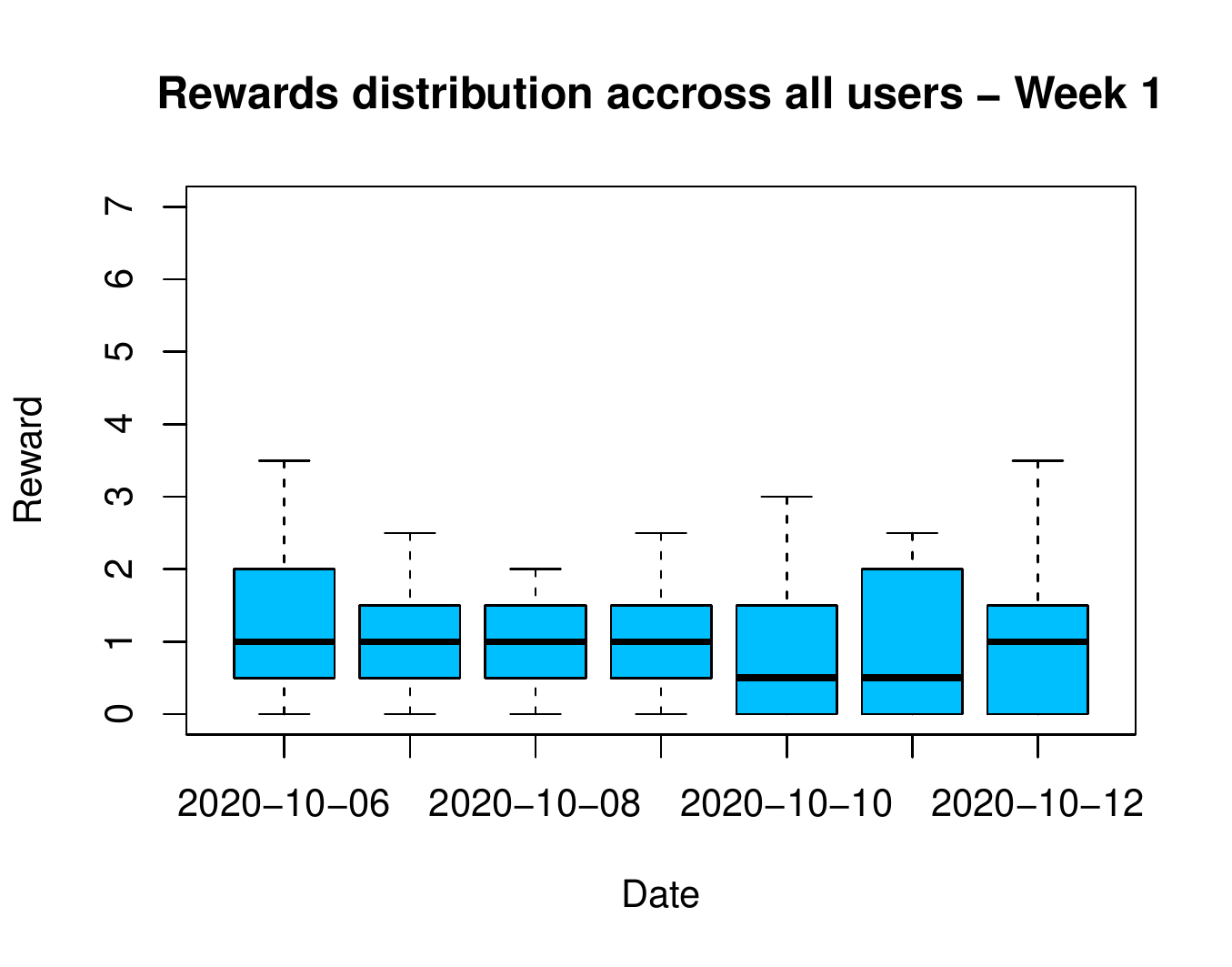}&
    \includegraphics[width=\textwidth]{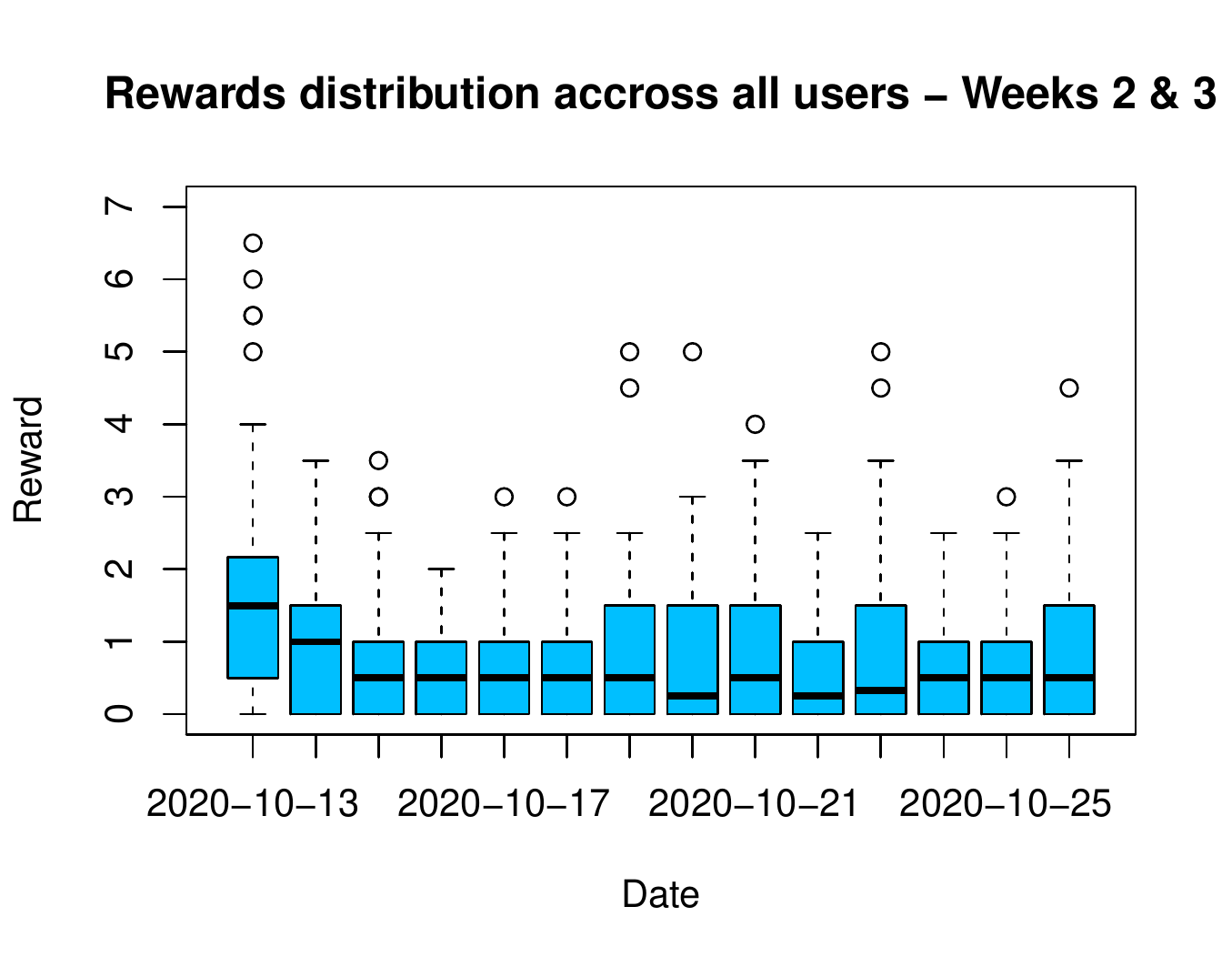}&
    \includegraphics[width=\textwidth]{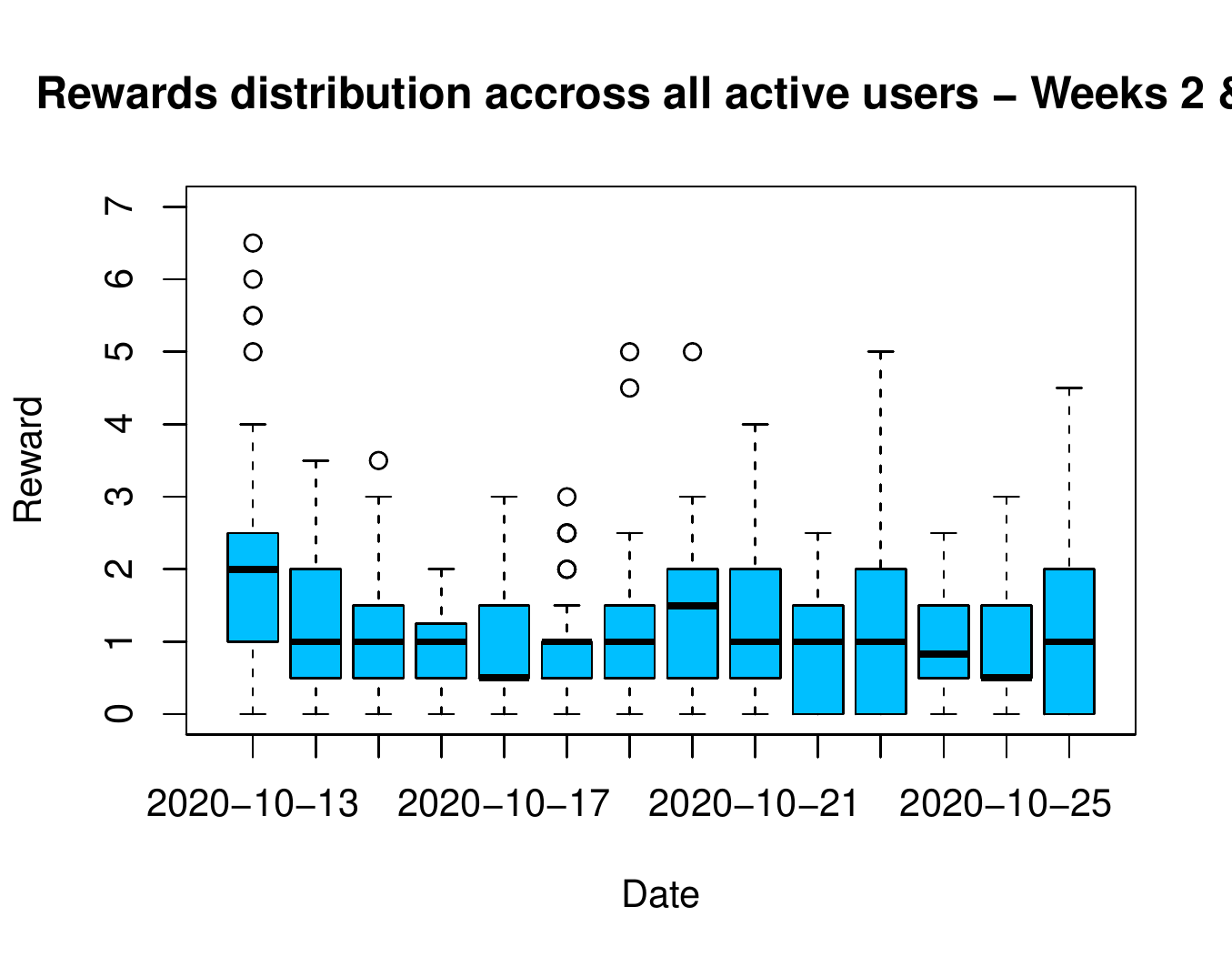}
    \end{tabular}
    \end{adjustbox}
    \caption{The box plots of the average daily reward per day part for weeks 1, 2 and 3.}
\label{figure3}
\end{figure}

\section{Discussion}
This paper presented a general reinforcement learning architecture for personalization in e-Health mobile applications (pH-RL). This architecture utilizes traces of states and rewards obtained from users to form clusters with k-means and the Dynamic Time Warping distance function. We built and integrated our architecture with the MoodBuster application for mood support. We ran an experiment with human participants for three weeks. Our results show that the pH-RL architecture can learn policies that consistently converge and provide the users with the right actions. Based on the observed reward values, we can conclude that pH-RL leads to increased adherence to the MoodBuster application. We ran our experiment for a period of two weeks. This resulted in a significant amount of user experiences. Given the number of users, we opted for training one policy across these users. In future work, it would be of great importance to experiment with a larger number of participants to find large enough clusters. Furthermore, the experiment can be ran for a longer period of time to evaluate the behaviour of the architecture during a longer period of time. Finally, including user specific features that describe their characteristics and preferences to the state space could result in better performance of the policies.

\newpage
\section{Appendix}

We use personalized motivational feedback messages to improve adherence to an online course for low mood. We define three groups of messages inspired by \cite{mol2018behind}. 

\subsection{Encouraging}
\subsubsection{Positive neutral mood}
\begin{itemize}
  \item It seems like you’re on the right track! Keep up the good work!
  \item Good to see that you are doing well. Good luck continuing Moodbuster Lite.
  \item You are making a lot of progress! You can be proud of yourself!
\end{itemize}

\subsubsection{Negative neutral mood}
\begin{itemize}
 \item It is good that you take part in Moodbuster Lite. You can commend yourself for that!
 \item Good that you are still rating your mood on a regular basis! Keep up the good work!
 \item It’s great that you are making time for yourself to improve your mood!
\end{itemize}

\subsubsection{Mood unavailable}
\begin{itemize}
  \item It may sometimes be difficult to engage in a training like Moodbuster Lite, but you can do it!
  \item Good that you started with Moodbuster Lite this is already a first step.
  \item It may be difficult to always keep the training and mood ratings on mind, but it’s great that you already started.
  \item Don’t give up if you haven’t rated your mood.
\end{itemize}

\subsection{Informing}
\begin{itemize}
  \item Don’t forget to set a reminder in order to not forget about your scheduled pleasant activities.
  \item Do you know you can always review the material of sessions if you need it?
  \item In your calendar, you can see the activities which you planned in the past.
  \item It is good to track what pleasant activities you did.
  \item Do not forget to rate your mood three times per day.
  \item Did you forget what pleasant activities to do? You can always check your notes on the website.
  \item It is sometimes helpful to re-read the content of the training to refresh your knowledge.
  \item Reminders may help you to not forget about the pleasant activities.
  \item It’s good to keep track of what pleasant activities you do and how your mood is.
\end{itemize}

\subsection{Affirming}
\subsubsection{Positive neutral mood}
\begin{itemize}
  \item Good to see that you are doing well.
  \item Even if you don’t rate your mood at some point, don’t worry, it may be hard to always think about it.
  \item I am happy to see that you feel well.
\end{itemize}

\subsubsection{Negative neutral mood}
\begin{itemize}
  \item It is very common to sometimes have low mood, so do not worry if that happens.
  \item It may be difficult to always keep the training and mood ratings on mind, but it is important for your well-being.
  \item Many people sometimes feel sad, this is nothing to worry about.
  \item It must be complicated to engage in this training, so it’s completely fine if you sometimes feel that way.
  \item Struggling to find the time to do the scheduled pleasant activities is completely normal, so do not worry if it ever happens to you.
\end{itemize} 
      
\subsubsection{Mood unavailable}
\begin{itemize}
  \item Don’t get discouraged if you forget to rate your mood sometimes, it’s normal. 
  \item It is completely normal to sometimes feel demotivated. 
  \item If you keep forgetting to do the pleasant activities? It will be ok! Don’t give up! 
  \item Do you often feel tired? That can easily happen when doing a training like this! \item It can be hard to think constantly about rating your mood.
  \item Struggling to find the time to do the scheduled pleasant activities is completely normal, so do not worry if it ever happens to you.
  \item It must be complicated to engage in this training, so it’s completely fine if you sometimes feel that way.
\end{itemize}    
    






\bibliographystyle{splncs04}
\bibliography{ijcai20}

\end{document}